\documentclass[conference]{IEEEtran}
\usepackage{grffile}
\usepackage{color}
\usepackage{graphicx}
\usepackage{multicol}
\usepackage{subcaption}
\usepackage{algcompatible}
\usepackage{algpseudocode}
\usepackage{siunitx}
\usepackage{amsmath}
\usepackage{array}
\usepackage{caption}
\usepackage[linesnumbered,algoruled,boxed,lined]{algorithm2e}
\usepackage{lipsum,graphicx,subcaption}
\usepackage{comment}
\usepackage{romannum}
\usepackage{float}

\begin{document}

\textcopyright 2019 IEEE. Personal use of this material is permitted.  Permission from IEEE must be obtained for all other uses, in any current or future media, including reprinting/republishing this material for advertising or promotional purposes, creating new collective works, for resale or redistribution to servers or lists, or reuse of any copyrighted component of this work in other works.

\newpage

\title{A Ride-Matching Strategy For Large Scale Dynamic Ridesharing Services Based on Polar Coordinates\\}
\author{\IEEEauthorblockN{Jiyao Li, Vicki H. Allan}
\IEEEauthorblockA{Department of Computer Science, Utah State University, Logan, Utah, 84322\\
Email: jiyao.li@aggiemail.usu.edu, vicki.allan@usu.edu}
}
\maketitle

\begin{abstract}
In this paper, we study a challenging problem of how to pool multiple ride-share trip requests in real time under an uncertain environment. The goals are better performance metrics of efficiency and acceptable satisfaction of riders. To solve the problem effectively, an objective function that compromises the benefits and losses of dynamic ridesharing service is proposed. The Polar Coordinates based Ride-Matching strategy (PCRM) that can adapt to the satisfaction of riders on board is also addressed. In the experiment, large scale data sets from New York City (NYC) are applied. We do a case study to identify the best set of parameters of the dynamic ridesharing service with a training set of 135,252 trip requests. In addition, we also use a testing set containing 427,799 trip requests and two state-of-the-art approaches as baselines to estimate the effectiveness of our method. The experimental results show that on average 38\% of traveling distance can be saved, nearly 100\% of passengers can be served and each rider only spends an additional 3.8 minutes in ridesharing trips compared to single rider service.
\end{abstract}


\section{INTRODUCTION}
In recent years, dynamic ridesharing is a popular service across the world. However, dynamic ridesharing is a challenging problem. First, trip requests are not known in advance. Thus, the environment is uncertain. Secondly, decisions have to be made in real time for matching a large number of drivers and riders, otherwise, riders will give up requesting and switch to alternative services. Thirdly, a dynamic ridesharing service has benefits that it makes trips more effective while it also causes inconvenience to riders (as riders have to spend more trip time). Therefore, the trade-off between benefits and losses must be taken into account.

In practice, many papers focusing on various versions of dynamic ridesharing have been published \cite{shen2016dynamic}. Yet, previous studies have the following limitations: $(\romannum{1})$ Riders are forced to specify constraints like earliest departure time or/and latest arrival time such that optimization can be done based on the constraints. However, in most cases, passengers have no exact idea on the constraints or are able to relax constraints within an acceptable range in exchange for lowering the fare. Hence, it is  hard for them to confirm exact values of constraints. $(\romannum{2})$ Most studies have not considered how to balance the trade-off between benefits and losses in the ridesharing system. Most existing proposals only consider objectives in one aspect (only maximize the benefit aspects or minimize the loss aspect), but never try to combine the benefit and loss together. $(\romannum{3})$ Most existing work uses fake data or small set of real data to verify their approach. Such results make it hard to estimate the effectiveness of their methods in a real world situation. 

To address the limitations of previous works, we make the contributions as follows:
\begin{itemize}
    \item We create a Unified Index (UI) estimating performance of dynamic ridesharing service by a  parameterized objective function. The optimization of the objective function does not rely on users specifying constraints, freeing riders from inputting constraints (like specifying they prefer the latest arrival time).
    \item We propose a Polar-Coordinates based Ride-Matching strategy (PCRM) to pool multiple riders into a vehicle. Polar coordinates are used to define a Spatially Constrained Zone (SCZ) along the route of a vehicle. In addition, the SCZs will shrink adaptively to balance benefit and loss of the dynamic ridesharing service. 
    \item We use large scale data sets of NYC to evaluate our method. The empirical results show that about 38\% of the traveling distance can be saved and nearly 100\% of passengers can be served by the service.
\end{itemize}


\section{Related Work}
\label{s2}
In this section, we highlight the most relevant works on dynamic ridesharing service in terms of techniques that are applied.

Different kinds of data structures for solving ridesharing have been studied. \cite{ma2013t} proposed a spatio-temporal index to store information on passengers and taxis. Although the speed of ride-matching can be accelerated, much computational cost has to be spent on information updating. \cite{huang2014large} introduced a kinetic tree method that stores all possible routes in an effective way and facilitates searching the best route at any time. However, passengers are forced to input time constraints, on which the tree insertion relies. 

Various search techniques have also been applied in previous works. \cite{coslovich2006two} proposed a 2-opt perturbation heuristic in the route neighborhood which is adapted to maximize passenger satisfaction. A genetic algorithm was applied by \cite{herbawi2012genetic} to improve the average serving rate and reduce the total traveling distance of vehicles. \cite{santos2013dynamic} introduced local search to seek for a sub-optimal route such that improving the serving rate and reducing distance can both be achieved. The drawback of these three papers is that authors used a local search or a genetic algorithm such that the dispatch platform fails to make a quick response to passengers in a real time situation. \cite{pelzer2015partition} aims to maximize mileage saving and ridesharing potential by partition-based ride-matching, yet the satisfaction of riders is ignored. \cite{chen2017data} utilized the source and destination of the primary rider to find appropriate other riders.  However, it did not consider location in the route, and the constraint search area cannot adapt to the environment.

Moreover, mechanism design has been introduced to the field of ridesharing as well. \cite{kamar2009collaboration} used the VCG auction \cite{wooldridge2009introduction} to distribute cost among riders so that a rider can pay less and incentive compatibility can be achieved. \cite{kleiner2011mechanism} proposed a second price sealed bid auction \cite{wooldridge2009introduction} to request riders’ and drivers’ preferences such that total distance can be minimized and serving rate can be maximized. However, it assumes that one driver can only pick up one rider, which limits flexibility. \cite{shen2016online} designed an online greedy approach to reduce the operation cost (total distance) of the whole system such that passengers' fare can be decreased. However, the greedy approach is unstable and it could be worse in some situations.

\section{Dynamic Ridesharing Problem}
\label{s3}
\subsection{Preliminary}
\label{preliminary}
A passenger is affected by the following measures.

\textbf{Definition 1: Waiting Time.} The waiting time of a trip request $T_w^r$ is defined as the time interval between when the trip request is accepted and the rider is picked up, where request $r \in R^+$.

We denotes $R^+$ as all the served trip requests and $R^-$ as all the rejected trip requests. We further denote $R=R^++R^-$ as the set of all trip requests in the dynamic ridesharing service.

\textbf{Definition 2: Trip Time.} The trip time of a trip request $T_t^r$ is defined as the time interval between when the rider is picked up and when the rider has been dropped off, where $r \in R^+$.

\textbf{Definition 3: Inconvenient Index (ICI).} The inconvenient index of a trip request $ICI^r$ indicates extra time that a passenger has to spend in completing their trip.  It is modeled as the time difference between ridesharing and individual driving.   We make the assumption that a passenger's inconvenience in the dynamic ridesharing service  is defined solely by the extra time that he or she spends on the journey.

\begin{equation}
ICI^r = C_w \cdot T_w^r + C_t \cdot (T_t^r - T_s^r)
\label{ICI}
\end{equation}

As shown in Eq.\ref{ICI}, where $C_w$ and $C_t$ are constant parameters indicating the penalty associated with long waiting time and extra trip time, $T_s^r$ is the time used by a rider if he or she drives individually.  This value can be estimated by the distance from the origin to the destination. We may further define the inconvenient index of the platform as Eq.\ref{ICI_1} shown.

\begin{equation}
ICI(R) = \sum \limits_{r \in R^+}\big(C_w \cdot T_w^r + C_t \cdot (T_t^r - T_s^r)\big)
\label{ICI_1}
\end{equation}

Secondly, indexes that measure the benefits of the platform are defined below. Here, we only consider the mileage saving index and the serving ability index. 

\textbf{Definition 4: Mileage Saving Index (MSI).} The mileage saving index of the platform indicates the percentage of the total mileage saved by dynamic ridesharing service. It is defined as the mileage difference between individual and ridesharing driving compared to the total mileage traveled in ridesharing.

\begin{equation}
MSI(V) = \frac{\sum \limits_{v \in V}(M_{single}^v-M_{share}^v)}{\sum \limits_{v \in V}M_{share}^v}
\label{MSI}
\end{equation}

As shown in Eq.\ref{MSI}, where $V$ is the set of vehicles that participate in the dynamic ridesharing service, $M_{single}^v$ is the sum of single driving distances of all trip requests served by vehicle $v$, which can be estimated by the sum of distances of origins and destinations, and $M_{share}^v$ represents the mileage traveled by vehicle $v$ for the ridesharing service.    

\textbf{Definition 5: Serving Ability Index (SAI).} The serving ability index reflects the amount of passengers that the platform is able to serve. It is defined as the number of trip requests that has been served $|R^+|$ over the the number of trip requests made to the ridesharing service $|R|$.
\begin{equation}
SAI(R)=\frac{|R^+|}{|R|}
\label{SAI}
\end{equation}

\subsection{Problem Formulation}
\label{prob_formu}
Basically, the Dynamic Ridesharing Problem (DRSP) is a NP-hard problem \cite{santos2013dynamic} and is defined as follows: given that a fleet of vehicles travel in an uncertain environment where trip requests of passengers emerge unexpectedly and will disappear in a short time period, we aim to find  an approach such that the benefits of the platform can be maximized, and at the same time, the inconvenience of passengers can be minimized.

From the previous section, each index is correlated with each other. For example, improving SAI might be at the expense of ICI and so forth. In order to optimize possible conflicting indexes in a balanced way, a multi-objective function is created. We attempt to unify all optimization indexes to create a Unified Index (UI) measuring the performance of the platform through the objective function. 

\textbf{Definition 6: DRSP.} Given a set of vehicles $V$ ($v \in V$), a set of trip requests $R$ where life cycle of $r$ ($r \in R$) is $[r.t, r.t+r.p]$, the DRSP is to match $v$ and $r$ into a pair $(v,r)$ and to find a route for each vehicle $v$, such that the Unified Index $UI(V,R)$ is maximized.

\begin{equation}
UI(V,R) = \alpha \cdot MSI(V) + \beta \cdot SAI(R) - \gamma \cdot ICI(R)
\label{UI}
\end{equation}

where $\alpha$, $\beta$ and $\gamma$ define the relative importance of the different indexes. The contribution due to ICI is subtracted as lower ICI is better.   The objective function  offers the flexibility to adjust the optimization goals for specific applications. For example, when $\alpha=0$ and $\gamma=0$, it will become a problem that maximizes the number of served trip requests; when $\gamma=0$ and $SAI(R)=1$, then it will become a problem that maximizes mileage saving; when $\alpha=0$ and $SAI(R)=1$, then it will be a problem minimizing inconvenience of passengers.

\section{Method}
\label{s4}
We propose the PCRM strategy, which considers the proximal and directional areas along with the route of a vehicle in determining which additional requests to consider. Both source and destination of accepted trip requests must be located inside the Spatially Constrained Zone (SCZ). In our strategy, we consider sources first before considering the associated destinations.

\subsection{Source-Side Ride-Matching}
\begin{figure}[!h]
\centering
\subcaptionbox{Source-Side}[.45\linewidth][c]{\includegraphics[width=.3\linewidth]{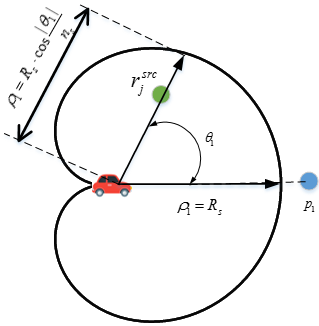}} \quad
\subcaptionbox{Destination-Side}[.45\linewidth][c]{\includegraphics[width=.5\linewidth]{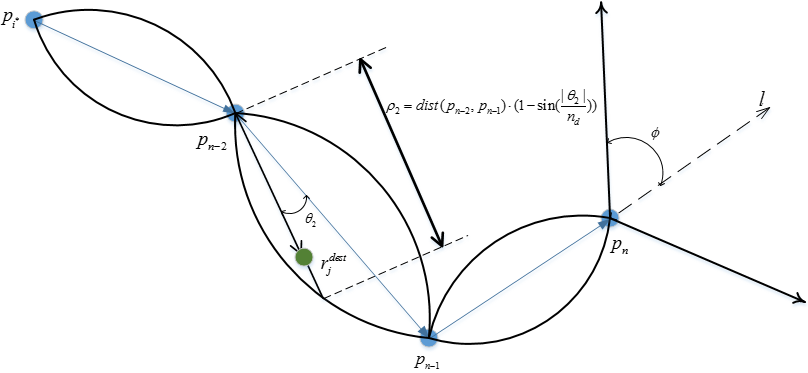}}
\caption{Ride-Matching Strategy}
\label{src_dest_match}
\end{figure}

The source-side ride-matching determines whether the source of a trip request is located in the SCZ. If so, the trip request will enter into the next phase where its destination will be evaluated; otherwise, it will not be considered further. The SCZ of the source side is drawn as shown in Fig.\ref{src_dest_match}(a). In the figure, $p_1$ is the first point in the route of a vehicle, in other words, it is the location that the vehicle is heading towards. The SCZ is defined by a polar coordinate system: the location of the vehicle serves as pole, the ray of the vehicle towards $p_1$ is regarded as polar axis, $\rho_1$ is polar radius and $\theta_1$ is polar angle ($-\pi \leq \theta_1 \leq \pi$). Inequalities (\ref{source_side_degree}) and   (\ref{source_side_ineq}) specify whether the source of a trip request is located inside the SCZ, where $dist(v_i, r^{src}_j)$ is the straight line distance between a vehicle $v_i$ and the source of the trip request $r^{src}_j$, $\theta_1$ is angular degree between direction of $v_i$ towards $r^{src}_j$ and the polar axis, $n_s$ is an adjustment factor of $\theta_1$ and its range is $0<n_s \leq 2$. We observe that source $r^{src}_j$ has more chance of being within the SCZ when it is in or near vehicle's heading direction ($|\theta_1|$ gets smaller); if it gets far from vehicle's heading direction ($|\theta_1|$ gets larger), there will be less chance of being in the zone unless it is very close to the vehicle.  

\begin{align}
0 \leq &\frac{|\theta_1|}{n_s} \leq \frac{\pi}{2} \label{source_side_degree} \\
dist(v_i, r^{src}_j) &\leq R_s \cdot \cos \frac{|\theta_1|}{n_s}
\label{source_side_ineq}    
\end{align}

If the source of a trip request is within the SCZ of source side, then we will use insertion algorithm \cite{tong2018unified} to find the appropriate position for $r^{src}_j$ in the route of the existing vehicle. The computational complexity of the insertion is linear.

\subsection{Destination-Side Ride-Matching}
After inserting $r^{src}_j$ as $p_{i^\ast}$ into the route of one vehicle, we will verify whether the destination $r^{dest}_j$ of a trip request $r_j$ is inside the SCZ of destination side. Both source and target of the trip request must be acceptable. As destination $r^{dest}_j$ must be inserted after its source $r^{src}_j$ in the route, so we only consider points after $p_{i^\ast}$ in the route (includes $p_{i^\ast}$). The SCZ of the destination side is shown in Fig.\ref{src_dest_match}(b). There are two types of SCZ in the destination side: one is the oval shaped zones among points along the route, the other is the triangular shaped zone at the tail of the route.

The oval shaped zones of the destination side starts from $p_{i^\ast}$ to the last point $p_n$ of the route, along with the trajectory of the route of vehicle. In the Fig.\ref{src_dest_match}(b), the oval shaped zones are also defined using polar coordinates: points of the route serve as poles (e.g. $p_{n-2}$), the ray of a point towards its next point in the route is regarded as polar axis, $\rho_2$ is the polar radius and $\theta_2$ is polar angle ($-\frac{\pi}{2} \leq \theta_2 \leq \frac{\pi}{2}$). Inequalities (\ref{dest_side_degree}) and (\ref{dest_side_ineq}) specify whether the destination of a trip request $r^{dest}_j$ is located inside an oval zone, where $dist(p_i, r^{dest}_j)$ is the straight line distance between a route point $p_i$ and the source of the trip request $r^{dest}_j$, $dist(p_i, p_{i+1})$ is the straight line distance between two adjacent points of the route of a vehicle, $n_d$ is an adjustment factor of $\theta_2$ and its range is $0<n_d \leq 1$.      

\begin{align}
0 \leq &\frac{|\theta_2|}{n_d} \leq \frac{\pi}{2} \label{dest_side_degree} \\
dist(p_i, r^{dest}_j) \leq &dist(p_i, p_{i+1}) \cdot (1-\sin{\frac{|\theta_2|}{n_d}})
\label{dest_side_ineq}
\end{align}

The triangular shaped zone is basically the extension of the last oval zone to increase pooling probability in the same direction, as shown in Fig.\ref{src_dest_match}(b). After the last point in the route, no other riders are affected by future drop off points.  Inequality (\ref{dest_side_ineq2}) indicates whether the destination of a trip request $r^{dest}_j$ is located  inside in the triangle shaped zones, where $\ell$ is the extension of $p_{n-1}p_n$ and $\phi$ is the angle threshold of the zone .
\begin{align}
angle(\overrightarrow{\rm p_{n}r^{dest}_j}, \overrightarrow{\rm \ell}) \leq \phi
\label{dest_side_ineq2}
\end{align}

In addition, if $r^{dest}_j$ is located inside the oval shaped zone of which the pole is $p_i$, then $r^{dest}_j$ will be directly inserted in the position between $p_i$ and $p_{i+1}$ in the route. If $r^{dest}_j$ is located  inside the triangular shaped zone, then $r^{dest}_j$ will be inserted at the tail of the route. In order to improve the probability of arrival pickup and delivery point on time, routing technique in \cite{cao2016multiagent} can be applied to select optimal path between points in the route.

\subsection{Adaptive SCZ}
Generally speaking, the Unified Index (UI) is affected by the area of SCZ of a vehicle. For example, if the zone area gets larger, more riders can be served but riders will suffer from longer delayed time. Our purpose is to maximize the UI such that the benefits and drawbacks of the dynamic ridesharing service can be balanced. We observe that the polar radius determines the area of SCZ. Basically, if the polar radius is increased, the area of SCZ will be larger. The serving rate will be improved but the inconvenience index will also be increased and vice versa. Hence, our basic idea is to adapt the length of polar radius to the ICI of riders on board: when the ICI gets larger, the polar radius of the SCZ will shrink adaptively; when the ICI gets smaller, polar radius will be longer and if the ICI is 0, polar radius will revert back to the initial length. As shown in Eq.(\ref{shrink}), $T_{lw}^{\hat{r}}$ and $T_{lt}^{\hat{r}}$ are the longest waiting time and trip time of riders currently in a vehicle, $n$ represents either the value of $n_s$ or $n_d$, $N$ stands for either the value of $N_s$ or $N_d$ ($N_s$ is the initial value of $n_s$ and $N_d$ is the initial value of $n_d$), the $\tau$ is the adaptive rate.      

\begin{equation}
n=N- (e^{\frac{C_w \cdot T_{lw}^{\hat{r}} + C_t \cdot \max(0,T_{lt}^{\hat{r}} -T_{s}^{\hat{r}})}{\tau}}-1)
\label{shrink}
\end{equation}

\section{Experiment}
\label{s5}
The experiment section is composed of two parts. In the first part, we do a case study of ridesharing in NYC based on taxi trip requests during peak hours of a weekday, where a set of parameters that fits the PCRM well is identified. In the second part, we compare our method to other state-of-art approaches based on a large scale taxi trip requests of NYC on another weekday. For simplicity, we assume that the capacity $ca$ of a vehicle is 4, the patience period $p$ of each rider is 20 min, and Manhattan distance \cite{nerbonne1997measuring} is used to measure the distance between two pair of points. Using vehicle specific and rider specific parameters is not expected to affect the results. All the values of constant parameters for DRSP and PCRM are shown in Table \ref{table1}. All the simulations are implemented by Python 3.5 and executed by a machine with Intel Corei7-3770 CPU (3.4GHz, quad-core) with 16GB memory.

\begin{table}[h]
\captionsetup{skip=0pt}
\caption{List of Constant Parameters for DRSP and PCRM} 
\begin{center}
\begin{tabular}{|c|c|c|}
\hline
Symbol & Definition & Value\\ \hline
$C_w$ & Penalty for waiting time & 1.1  \\ \hline
$C_t$ & Penalty for extra trip time & 1  \\ \hline
$\alpha$ & The relative importance of MSI & 1 \\ \hline
$\beta$ & The relative importance of SAI & 1 \\ \hline
$\gamma$ & The relative importance of ICI & 0.1 \\ \hline
$N_s$ & The initial value of $n_s$ & 2 \\ \hline
$N_d$ & The initial value of $n_d$ & 1  \\ \hline
\end{tabular}
\end{center}
\label{table1}
\end{table}

\subsection{A Case Study of NYC}
The time period of trip requests used in the training set is from 6PM to 12AM of a weekday. There are a total of 135,252 trip requests generated during the time period and 6000 vehicles are used. 

There are 5 parameters that can be tuned in the PCRM: the polar radius $R_s$, the directional angle $\phi$, the adaptive rate $\tau$, the polar angle adjustment factors $n_s$ and $n_d$. Since $n_s$ and $n_d$ can be adjusted automatically adapting to the ICI of riders, and the value of $\tau$ is set to 20 by experimental tuning, we focus on considering the parameters $R_s$, $\phi$.

\begin{figure}[!h]
  \centering
  \subcaptionbox{ICI}[.45\linewidth][c]{\includegraphics[width=.4\linewidth]{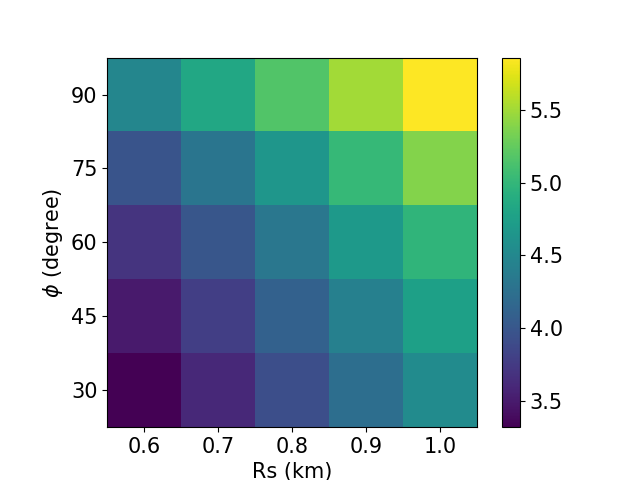}}\quad
  \subcaptionbox{MSI}[.45\linewidth][c]{\includegraphics[width=.4\linewidth]{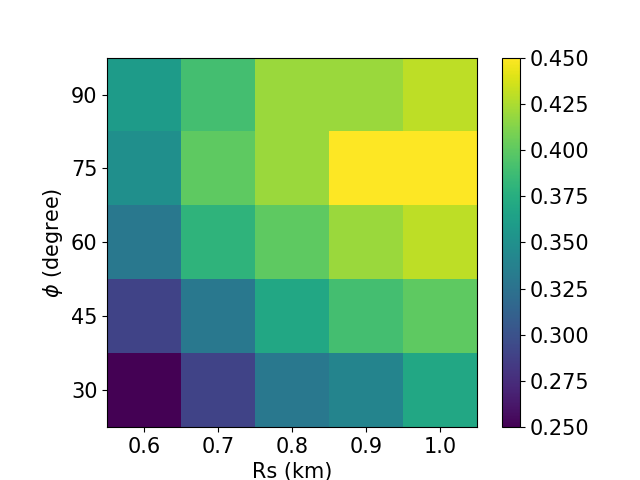}}
  \caption{Simulation results for the ICI and MSI on $R_s$ and $\phi$. Low values of ICI are better, while high values of MSI are better.}
  \label{ici-msi-heat}
\end{figure}

In Fig.\ref{ici-msi-heat}(a), the values of ICI increase as $R_s$ and $\phi$ increase.  In other words, riders will feel more uncomfortable as $R_s$ and $\phi$ get larger. This is because drivers will have more chance to pick riders up such that riders on board will have an increased trip length. In Fig.\ref{ici-msi-heat}(b), the MSI increases along with the increase of $R_s$ and $\phi$ when $\phi \in [45,75]$. However, the MSI decreases as $\phi$ gets larger when $\phi \in [75,90]$.  

\begin{figure}[!h]
  \centering
  \subcaptionbox{SAI}[.45\linewidth][c]{\includegraphics[width=.45\linewidth]{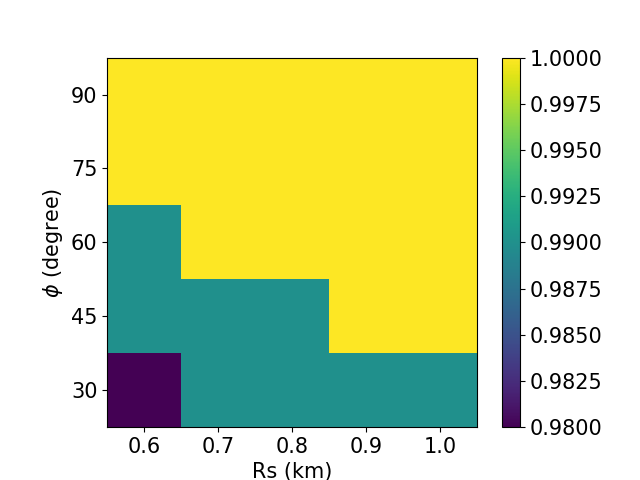}}\quad
  \subcaptionbox{UI}[.45\linewidth][c]{\includegraphics[width=.45\linewidth]{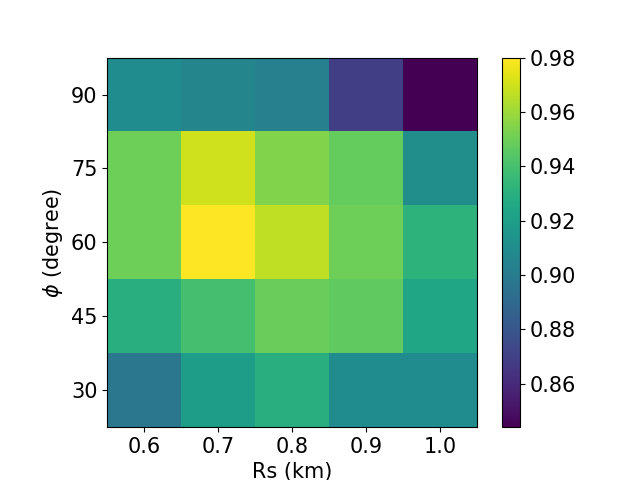}}
  \caption{Simulation results for the SAI and UI depend on $R_s$ and $\phi$}
  \label{sai-ui-heat}
\end{figure}

We are also concerned about the SAI, as shown in Fig.\ref{sai-ui-heat}(a).  We observe that SAI can reach $100\%$ if $R_s$ greater than 0.8 or $\phi$ larger than 45. Fig.\ref{sai-ui-heat}(b) shows UI along with different combinations of $R_s$ and $\phi$. We observe that the highest value of UI is about 0.97 with $R_s=0.7$ and $\phi=60$.

\begin{figure*}[!h]
  \centering
  \subcaptionbox{UI}[.2\linewidth][c]{\includegraphics[width=.23\linewidth]{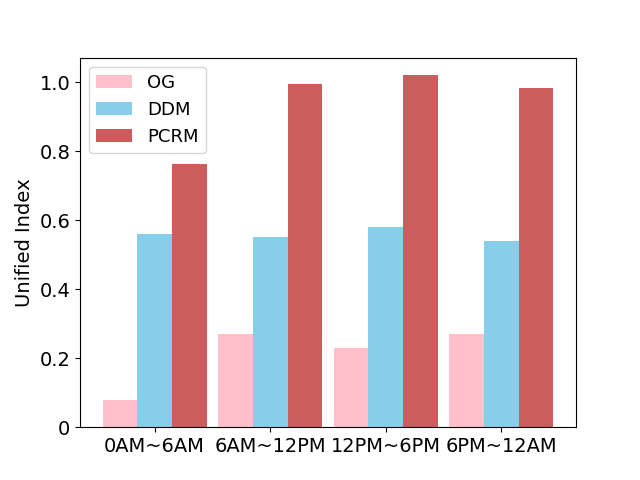}}\quad
  \subcaptionbox{SAI}[.2\linewidth][c]{\includegraphics[width=.23\linewidth]{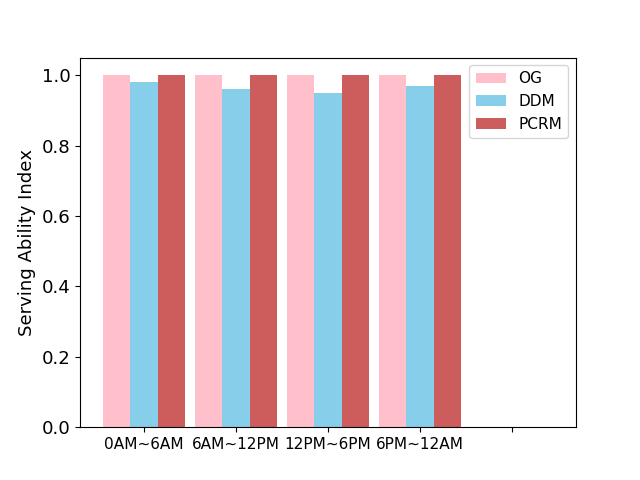}}\quad
  \subcaptionbox{MSI}[.2\linewidth][c]{\includegraphics[width=.23\linewidth]{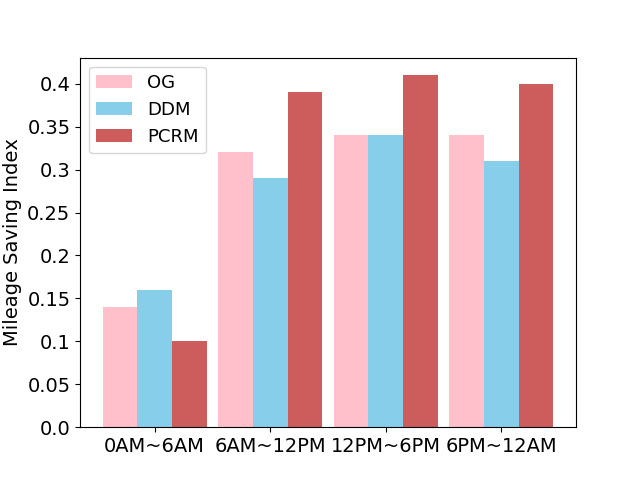}}\quad
  \subcaptionbox{ICI}[.2\linewidth][c]{\includegraphics[width=.23\linewidth]{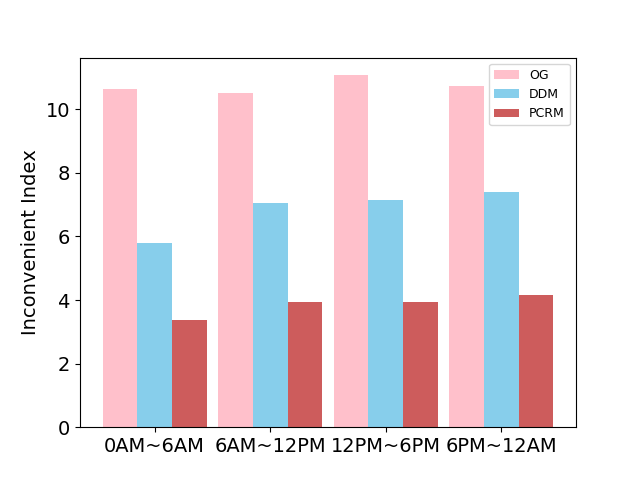}}
  \caption{Comparisons of the performance metrics among three approaches: PCRM, DDM and OG}
  \label{compare1}
\end{figure*}


\subsection{Comparison}
For evaluation purposes, we implement two other state-of-the-art algorithms to estimate the efficiency of the PCRM. The first algorithm used the data-driven method (DDM) to study the potential of ridesharing, the strategy of it is to select riders whose pickup location is inside the trapezoid region created by vehicle's heading direction, then applied DDM to find out the appropriate set of parameters of the strategy such that the all over efficiency(MSI, SAI and ICI) can be optimized., it has been proven to be efficient with several data sets of big cities \cite{chen2017data}. The other algorithm uses Online Greedy (OG) method to select a rider candidate who has smallest extra cost (extra cost means the extra distance used to pick up and deliver the new  rider compared to the original route), such that the average fare amount of riders in the vehicle can be minimized. This solution works well in the grid city simulation with random generated trip requests \cite{shen2016online}. Also, to evaluate the performance of the three algorithms, we use a testing set of 427,799 trip requests over a whole weekday in NYC (the data set used here is different from the data set in case study) and 7000 vehicles are deployed for the dynamic ridesharing service. 

We compute the UI that measures the compromising ability of dynamic ridesharing service using the following methods: PCRM, DDM and OG. As Fig.\ref{compare1}(a) shown, the results clearly show that the PCRM performs significantly better than others.The UI of the PCRM goes up to 0.99 while other two methods is 0.6 and 0.3 separately. This can be explained by the fact that the PCRM can adapt to the environment such that it can balance the benefits and losses well, yet the DDM and the OG are fixed methods such that they fail to adapt to different situations automatically.

 The SAI of the PCRM, DDM and OG is shown in Fig.\ref{compare1}(b), all three methods achieve nearly $100\%$. Fig.\ref{compare1}(c) shows the MSI with three methods. Between 0AM and 6AM, the MSI of all three methods is lowest at about $13\%$. This is because the trip request quantity is low and the spatial distribution of trip requests is sparse. The PCRM is slightly lower than other two methods since increasing MSI in low and sparse situations will result in higher ICI. After 6AM, the PCRM can save more mileage than the other two methods when the quantity and the distribution of trip requests grows and gets dense. The PCRM totally saves about 455,750 kilometers in the whole day. Fig.\ref{compare1}(d) shows the ICI with three methods. The PCRM is clearly the smallest in the whole day. This can be explained by the fact that the PCRM uses adaptive SCZ to address the satisfaction of riders on board.

In summary, the PCRM outperforms the DDM and the OG overwhelmingly in the performance metrics of ridesharing service. Although it still suffers from an extra 3.8 minutes compared with individual driving, the significant achievements in MSI and SAI can compensate for such trivial loss.

\section{Conclusion}
\label{s6}
In this work, we take the PCRM to study the performance metrics of large scale dynamic ridesharing service. At first, we define the performance metrics of the dynamic sharing service and formulate the dynamic ridesharing problem as an objective function that compromises the benefits and loss. Then we proposed the PCRM that pools appropriate riders into a single vehicle. At last, we applied large scale real world trip request data to do the experiment. The results show that about $38\%$ distance can be saved, nearly $100\%$ passengers can get the service and the compensation is only about 3.8 minutes delay (compare to drive individually) for each rider averagely. This also means the PCRM favor the city by reducing fuel consumption, $CO_2$ emission and the amount of vehicle on road.

In future work, we will apply real map to estimate the distance and reinforcement learning to learn the best parameters of the SCZ.

\bibliographystyle{plain}
\bibliography{jiyaoli-drs}

\end{document}